\documentclass[10pt,twocolumn,letterpaper]{article} %DO NOT CHANGE THIS

\usepackage{cvpr}  %Required
\usepackage{times}  %Required
\usepackage{epsfig}
\usepackage{graphicx}  %Required
\usepackage{amsmath}
\usepackage{amssymb}
\usepackage{helvet}  %Required
\usepackage{courier}  %Required
\usepackage{url}  %Required
\usepackage{color}
\usepackage{CJKutf8}%导入CJKutf8包，并且激活中文、日文、韩文的uft8编码

\usepackage{refstyle}
\usepackage{bm} % bold math symbols
\usepackage{subcaption}
\usepackage{appendix}
\usepackage[font={small}]{caption}
\usepackage[abs]{overpic}
\usepackage{dsfont}
\usepackage{multirow}
\usepackage[T1]{fontenc}
\usepackage[utf8]{inputenc}
\usepackage{authblk}
\usepackage{xcolor}
\usepackage{adjustbox}

\usepackage[utf8]{inputenc}
\usepackage{bibunits}%bib
\usepackage[breaklinks=true,bookmarks=false]{hyperref}  % change the previous line to this line for *******final submission*******
\usepackage[marginal]{footmisc}
\cvprfinalcopy % *** Uncomment this line for the *******final submission*******

\def\cvprPaperID{} % *** Enter the CVPR Paper ID here

\begin{document}

\title{SPI-Optimizer: an integral-Separated PI Controller for Stochastic Optimization}
\author{
%	% Institution1\\

	Dan Wang\textsuperscript{1$*$}, Mengqi Ji\textsuperscript{2}\thanks{\hangafter 1 
		\hangindent 1.8em equal contribution \\  wd17@mails.tsinghua.edu.cn\\ mji@connect.ust.hk} , Yong Wang\textsuperscript{1} , Haoqian Wang\textsuperscript{3} , Lu Fang\textsuperscript{1}\thanks{fanglu@sz.tsinghua.edu.cn} \\

	\textsuperscript{1} Tsinghua-Berkeley Shenzhen Institute, Tsinghua University, Shenzhen, China\\
	\textsuperscript{2} The Hong Kong University of Science and Technology, Hong Kong, China\\
	\textsuperscript{3} Graduate School at Shenzhen, Tsinghua University, Shenzhen, China\\

%	{\tt\small \textsuperscript{1}wd17@mails.tsinghua.edu.cn, \textsuperscript{2}mji@connect.ust.hk, \textsuperscript{1}fanglu@sz.tsinghua.edu.cn}
	% For a paper whose authors are all at the same institution,
	% omit the following lines up until the closing ``}''.
	% Additional authors and addresses can be added with ``\and'',
	% just like the second author.
	% To save space, use either the email address or home page, not both
	\and
%	 Second Author\\
	% Institution2\\
	% First line of institution2 address\\
	% {\tt\small secondauthor@i2.org}
}

%\thanks{mji@connect.ust.hk} \thanks{equal contribution} 

%\institute{\email{wd17@mails.tsinghua.edu.cn} \\Institute} %error, cvpr not
\maketitle
%\footnotetext[1]{These authors contributed equally to this work.} % To change the numbering, use the following command: the default is Arabic numerals\renewcommand{\thefootnote}{\alph{footnote}}  

%\let\thefootnote\relax\footnotetext{wd17@mails.tsinghua.edu.cn}
%\let\thefootnote\relax\footnotetext{mji@connect.ust.hk}

%\centerline{\large{Paper ID: 3170}}

%\begin{bibunit}%bib
% sections
\begin{abstract}

To overcome the oscillation problem in the classical momentum-based optimizer, recent work associates it with the proportional-integral (PI) controller, and artificially adds D term producing a PID controller. It suppresses oscillation with the sacrifice of introducing extra hyper-parameter.
In this paper, we start by analyzing: why momentum-based method oscillates about the optimal point? and answering that: the fluctuation problem relates to the lag effect of integral (I) term. Inspired by the conditional integration idea in classical control society, we propose SPI-Optimizer, an integral-Separated PI controller based optimizer WITHOUT introducing extra hyper-parameter. 
It separates momentum term adaptively when the inconsistency of current and historical gradient direction occurs. Extensive experiments demonstrate that SPI-Optimizer generalizes well on popular network architectures to eliminate the oscillation, and owns competitive performance with faster convergence speed (up to $40\%$ epochs reduction ratio ) and more accurate classification result on MNIST, CIFAR10, and CIFAR100 (up to $27.5\%$ error reduction ratio) than the state-of-the-art methods.

\end{abstract}
\section{Introduction}

% DL & optimizer

Serving as a fundamental tool to solve practical problems in both scientific and engineering domains, a proper optimizer plays vital role. Taking the highly concerned deep learning successful stories \cite{hinton2006reducing,huang2017densely,ji2017surfacenet,zheng2016deep,zagoruyko2016wide,he2016deep,huang2017densely} as examples, stochastic gradient descent (SGD) serves as one of the most popular solvers, due to its ability in maintaining a good balance between efficiency and effectiveness. The expectation in training very deep networks substantially requires for even more efficient optimizer such as SGD-Momentum (MOM) \cite{rumelhart1986learning}. However it suffers from the oscillation problem \cite{ogata1995discrete}, with non-negligible maximum overshoot and settling time. Such an oscillation phenomena hinders the
convergence of MOM, requiring more training
time and resources. As a result, an efficient as well as effective optimizer is urgently demanded yet very challenging, owing to the highly non-convex nature of the optimization problems.

Recently, some researchers investigate the conventional optimization problem by associating it with the Proportional-Integral-Derivative (PID) model that widely used in the feedback control system. By linking the calculation of errors in feedback control system and the calculation of gradient in network updating, \cite{an2018pid} shows that MOM can be treated as a special case of classical PID controller with only Proportional (P) and Integral (I) components. It further artificially adds the Derivative (D) component to form a PID based optimizer, which reduces the oscillation phenomena by introducing troublesome hyper-parameter induced by D component. In other words, the calculated coefficient of the derivative term can hardly adapt to the huge diversity of network architectures and different modalities of the training dataset.

On the contrary to extend PI to PID directly, we explore ``why momentum-based method oscillates about the optimal point?'' via thorough analysis from the perspective of inherent connection between MOM and PI controller. The in-depth pre-analysis (Section 3.1) reveals that the fluctuation problem in momentum-based method relates to the lag effect of integral (I) term in PI controller. Inspired by the conditional integration idea in classical control society, we propose SPI-Optimizer, an integral-Separated PI controller based solver for stochastic Optimization scheme.
SPI-Optimizer separates momentum term adaptively when the inconsistency of current and historical gradient direction occurs. 

More specifically, the insight of SPI-Optimizer can be explained more explicitly as follows (more discussions in Section 3.3). For Conditional Integration used in classical control society (denoted as CI), the integral component is only considered as long as the magnitude of the feedback deviation (the gradient) is smaller than a threshold $\beta$. That means SGD with only proportional (P) term can be viewed as CI-$\beta=0$. Similarly, MOM never separates out the integral (I) part and can be denoted as CI-$\beta=+\infty$. While the oscillation phenomenon may be tuned by setting $\beta$, the convergence speed of CI cannot be improved by trading off the parameter $\beta$, which remains bounded by CI-$\beta=0$ and CI-$\beta=+\infty$. 
Our SPI-Optimizer examines the sign consistency between the residual and the integral term before enabling the integral component, thus easing the oscillation phenomenon WITHOUT introducing extra hyper-parameter. As a result, it can be theoretically shown that SPI-Optimizer outperforms both CI-$\beta=0$ and CI-$\beta=+\infty$, owning more generalization ability for different network structures across several popular dataset. We summarize the technique contributions as follows.
\begin{itemize}
  \item By associating MOM with PI controller, we analytically show that the oscillation in momentum-based method corresponds to the lag effect of integral (I) term in PI controller, which inspires us to deal with I term instead of adding D term, as the latter one introduces extra hyper-parameter. 
  \item A novel SPI-Optimizer based on the integral-Separated PI controller is proposed to separate momentum term adaptively when the inconsistency of current and historical gradient direction occurs. The detailed discussion on the convergence of SPI-Optimizer is provided theoretically.
  \item SPI-Optimizer eliminates the oscillation phenomenon without introducing any extra hyper-parameter and leads to considerably faster convergence speed and more accurate result on popular network architectures.
\end{itemize}

% Extensive experiments demonstrate that SPI-Optimizer generalizes well on popular network architectures to eliminate the oscillation, and owns competitive performance with faster convergence speed (up to $35\%$ epochs reduction) and more accurate classification result (up to $40\%$ error reduction) on MNIST, CIFAR10, and CIFAR100.

%Mengqi's note:
%As same as the feedback system control, the coefficient of D component is too sensitive to be tuned.
%Because many factors, say the type of  it should be determined by the characteristics of the loss functions $L$ that depends on the training data $<x, y>$, the network structure $f(\cdot|\theta)$ and the metric between $y = f(x|\theta^*)$ and $y = f(x|\theta)$

\section{Related Work}

Among the various optimization schemes, the gradient based methods have served as the most popular optimizer to solve tremendous optimization problems. The representative ones include gradient descent (GD) \cite{cauchy1847methode}, stochastic gradient descent (SGD) \cite{robbins1951stochastic}, heavy ball (HB)\cite{polyak1964some}, Nesterov accelerated gradient (NAG) \cite{nesterov1983method} etc. While GD is the simplest one, it is restricted by the redundant computations for large dataset, as it recomputes gradients for similar examples before updating each parameter. SGD improves it by sampling a random subset of the overall dataset, yet it is difficult to pass ravines \cite{sutton1986two}. HB puts forward by adding a fraction to accelerate the iteration, which is further developed and named as Momentum(MOM) \cite{rumelhart1986learning}. NAG further uses the momentum term to update parameters and corrects gradient with some prescience. All of these classic optimization algorithms own fixed learning rates. 

%Conjugate Gradient \cite{wolfe1975method},
Lately, an increasing share of deep learning researchers train their models with adaptive learning rates methods \cite{duchi2011adaptive,zeiler2012adadelta,reddi2018convergence,kingma2014adam}, due to the requirement of speeding up the training time \cite{Andrej2017trends}. They try to adapt updates to each individual parameter to perform larger or smaller updates depending on their importance.

 Regardless the successful usage of adaptive method in in many applications owing to its competitive performance and its ability to work well despite minimal tuning, the recent findings by \cite{wilson2017marginal} show that hand-tuned SGD and MoM achieves better result at the same or even faster speed than adaptive method. Furthermore, the authors also show that for even simple quadratic problems, adaptive methods find solutions that can be orders-of-magnitude worse at generalization than
those found by SGD(M).
 They put forward that a possible explanation for the worse results in adaptive methods lies in the convergence to different local minimums \cite{im2016empirical}. It is also noted that most state-of-the-art deep models such as ResNet \cite{he2016deep} and DenseNet \cite{huang2017densely} are usually trained by momentum-based method, as the adaptive methods generalize worse than SGD-Momentum, even when these solutions have better training performance.

%To further improve SGD in terms of convergence rate and accuracy, researchers propose new methods based on stochastic variance reduced gradient (SVRG) \cite{huo2017asynchronous,liu2017accelerated,meng2017asynchronous}. SVRG usually uses some samples to estimate the gradient over a period of time, improving the convergence rate acceleration by variance reduction technique. However, the improvement degrades significantly when the dataset becomes sparse and large-scale. 

On the other hand, some researchers try to investigate stochastic optimization by associating it with the classical Proportional-Integral-Derivative (PID)  controller that widely used in the feedback control system. The pioneer work \cite{vitthal1995generalized} regarded the classical gradient descent algorithm as the PID controller that uses the Proportional (P) term merely. They added Integral (I) and Derivative (D) terms to achieve faster convergence. The latest work \cite{an2018pid} interpreted that momentum can be treated as a PI controller, and a Derivative (D) term that is the predictive gradient difference is added to reduce oscillation and improve SGD-Momentum on the large-scale dataset. Unfortunately, either introducing I and D terms to GD, or introducing D term to SGD-Momentum intensifies the task of tuning (which will be further elaborated in our experiments).

\section{SPI-Optimizer}
% In the network training process, the objective function is often stochastic and composed of a sum of subfunctions evaluated at different subsamples of data. An efficient optimization method is taking gradient steps w.r.t. individual subfunctions $g_t = \partial L_i / \partial w$ at $t$th iteration. Therefore, we

In this section, we firstly conduct a thorough pre-analysis on the oscillation phenomena of momentum-based algorithm in Section 3.1. Aided by the association with PI controller, the oscillation can be explained by the lag effect of integral (I) term  in  PI  controller. We then propose a novel SPI-Optimizer to separate I term from PI controller adaptively, which eases the oscillation problem WITHOUT introducing extra hyper-parameter. Subsequently, in-depth discussions to further evaluate SPI-Optimizer are provided in Section 3.3.

\subsection{Pre-analysis of Oscillation}
As introduced in \cite{rumelhart1986learning}, the Momentum algorithm (MOM) works by accumulating an exponentially
decayed moving average of past gradients. Mathematically, the momentum-based optimizer can be defined as
\begin{equation} \label{eq:MOM}
    \begin{aligned}
        v_{t+1} &= \alpha v_t + r \nabla L(\tilde{\theta}_t), \\
        \theta_{t+1} &= \theta_t - v_{t+1},
    \end{aligned}
\end{equation}
where $\tilde{\theta}_t = \theta_t$ and $\tilde{\theta}_t = \theta_t - \alpha v_t$ respectively define MOM and Nesterov Accelerated Gradient (NAG) \cite{nesterov1983method}.

\begin{figure}[ht!]
\centering

\newcommand{\colw}{0.48}
    \newcommand{\figw}{0.98} % can bigger than 1
    \begin{subfigure}[b]{\colw\linewidth}
        \includegraphics[width=\figw\textwidth,clip]{{{sections/figures/figure_files/T1_0_-2_1}}}
        \captionsetup{labelformat=empty}
        \caption{a: MOM's trajectory}
        \label{/fig:mom_lag_a}
    \end{subfigure}
    
    \begin{subfigure}[b]{\colw\linewidth}
        \includegraphics[width=\figw\textwidth,clip]{{{sections/figures/figure_files/overmom}}}
        \captionsetup{labelformat=empty}
        \caption{b: MOM}
        \label{/fig:mom_lag_mom}
    \end{subfigure}
        ~ 
    \begin{subfigure}[b]{\colw\linewidth}
        \includegraphics[width=\figw\textwidth,clip]{{{sections/figures/figure_files/oversgd}}}
        \captionsetup{labelformat=empty}
        \caption{c: GD}
        \label{/fig:mom_lag_gd}
    \end{subfigure}
        ~ 
    \begin{subfigure}[b]{\colw\linewidth}
        \includegraphics[width=\figw\textwidth,clip]{{{sections/figures/figure_files/overnag}}}
        \captionsetup{labelformat=empty}
        \caption{d: NAG}
        \label{/fig:mom_lag_nag}
    \end{subfigure}
        ~ 
    \begin{subfigure}[b]{\colw\linewidth}
        \includegraphics[width=\figw\textwidth,clip]{{{sections/figures/figure_files/overm99}}}
        \captionsetup{labelformat=empty}
        \caption{e: SPI}
        \label{/fig:mom_lag_isp}
    \end{subfigure}
\caption{(a): convergence path of MOM on a 2D toy convex function $f_1(\theta)$ with colored segments representing each weight update. (b-e): horizontal residual w.r.t. the optimal point versus time. Two components for the weight update are the current gradient (red arrow) and the momentum (green arrow), which can be interpreted as two forces dragging the blue curve.}
\label{/fig:mom_lag}
\end{figure}
% oscillation phenomenon
Although the momentum component can accelerate the convergence in the case of small and consistent gradients, it suffers from oscillation phenomenon that the convergence path fluctuates about the optimal point, as shown in Fig.~\ref{/fig:mom_lag_a}. Such oscillation can be quantitatively described by two concepts: the settling time $t_s$, defined as the time required for the curve to reach and stay within a range of certain threshold ($1e-2$) to the optimal point, and the maximum overshoot describing the difference between the maximum peak value and the optimal value:
\begin{equation} \label{eq:overshoot}
    \Delta\theta_{\max}^{(i)} = \theta_{\max}^{(i)} - \theta^{(i)*}
\end{equation}

% reason
As defined in Eqn. (\ref{eq:MOM}), there are two components contributing to the weight update, i.e., the momentum $-\alpha v_t$ and the current gradient $-r \nabla L(\theta_t)$. The integral term can introduce some non-negligible weight updates that are opposite to the gradient descent direction. In that case, the momentum will lag the update of weights even if the weights should change their gradient direction. Analogous to the feedback control, such lag effect leads to more severe oscillation phenomenon, i.e., the convergence path fluctuates about the optimal point with larger maximum overshoot $\Delta\theta_{\max}^{(i)}$ and longer settling time $t_s$.

We further take a commonly used function $f_1(\theta)={\theta^{(1)}}^2+50{\theta^{(2)}}^2$ for illustration.
In Fig.~\ref{/fig:mom_lag_a},
the convergence path is composed of multiple weight updates shown in different colors. By only considering the horizontal axis, Fig.~\ref{/fig:mom_lag_mom} depicts the residual of the convergence path to the optimal point using blue curve, and the weight updates from both the momentum $-\alpha v_t$ and the current gradient $-r \nabla L(\theta_t)$, shown as green and red arrows respectively.
In the process of approaching the goal, we define several time stamps: $t_1$ as the time when the curve first exceeds the optimal point, $t_2$ as the time when it reaches the maximum overshoot, and $t_s$ as the settling time.

The weight updates (green and red arrows) start with the same direction (up) until $t_1$. For the duration $[t_1, t_2]$, because the weight exceeds the optimal point (origin point in this specific example), the gradient descent direction (red arrow) gets reversed. But owing to the large accumulated gradients value (green arrow), the weight update deviates from the current rising trend until $t_2$ when $\alpha v_t + r \nabla L(\tilde{\theta}_t) = 0$.
As a result, the momentum introduces lag effect to the update of weights in the period of $[t_1, t_2]$ and leads to severe oscillation effect with large maximum overshoot and long settling time.

Compared with MOM, gradient descent (GD) oscillates less due to the lack of the accumulated gradients, shown in Fig.~\ref{/fig:mom_lag_gd}. Even though the maximum overshoot is much smaller than MOM, the settling time is unacceptably longer. Due to the lag effect of the momentum within the period $[t_1, t_2]$, the oscillation phenomenon of NAG is as severe as MOM. In Fig.~\ref{/fig:mom_lag}, SPI-Optimizer obtained about $55\% - 73\%$ convergent epoch reduction ratio.

\subsection{integral-Separated PI Controller} %\cite{sec_CI}
The latest work \cite{an2018pid} points out that if the optimization process is treated as a dynamic system, the optimizer Stochastic Gradient Descent (SGD) can be interpreted as a proportion (P) controller with $\alpha=0$.
Then, MOM and NAG can be represented as Proportional-Integral (PI) controller. As discussed in the previous subsection, the momentum term can lead to severe system oscillation about the optimal point owing to the lag effect of the integral / momentum. To ease the fluctuation phenomenon, \cite{an2018pid} artificially adds the derivative (D) term to be a PID-$K_d$ controller with a network-invariant and dataset-invariant coefficient $K_d$.
However, it is questionable that an universal (D) coefficient $K_d$, rather than a model-based one, is applicable to diverse network structures.
At the same time, the newly introduced hyper-parameter for the derivative term needs more effort for empirical tuning. In contrast, we propose integral-Separated PI Controller based Optimizer (SPI-Optimizer) to directly deal with the integral term, WITHOUT introducing any extra hyper-parameter.

% feedback control system objective: gradient --> 0
In a typical optimization problem, the loss function $L$ is a metric to measure the distance between the desired output $y$ and the prediction $f(x|\theta)$ given the weight $\theta$. The gradient of the weights $\nabla L(\theta)$ can be used to update the weights till the optimal solution with zero gradient.
Hence the gradient $\{\nabla L(\theta_i)\}_{i=0, \cdots, t}$ w.r.t. weights can be associated with the ``error'' in the feedback control. Consequently, rethinking the problem from the perspective of control, although PI controller leads to faster respond compared with P controller, it can easily lag and destabilize the dynamic system by accumulating large historical errors.

%Inspired by the conditional integration [XXXXXXX] strategy in control community, preventing the integral term from accumulating within pre-determined bound can  suppress the lag effect effectively.
%Analogously, we can simply define a Conditional Integration (CI) inspired optimizer as follows:
Inspired by the conditional integration \cite{astrom1989integrator} strategy in control community, which prevents the integral term from accumulating within pre-determined bound to effectively suppress the lag effect, a simple conditional integration optimizer (CI-$\beta$) is proposed as follows:
\begin{equation} \label{eq:CI}
    \begin{aligned} 
        %v_{t+1} &= \alpha v_t - r \nabla L(\tilde{\theta}_t) \\
        v_{t+1}^{(i)} &= \alpha v_t^{(i)} \mathds{1}\{|\nabla L(\theta_t^{(i)})| < \beta\} + r \nabla L(\theta_t^{(i)}) \\
        \theta_{t+1}^{(i)} &= \theta_t^{(i)} - v_{t+1}^{(i)},
    \end{aligned}
\end{equation}
 where $\beta$ is the introduced threshold for each dimension of the state vectors. Unfortunately, such naive adoption leads to some drawbacks: (1) it requires extra effort to empirically tune the hyper-parameter $\beta$, and $\beta$ has weak generalization capability across different cost function $L$, (2) by manually selecting the gradient threshold, the performance of CI-$\beta$ is almost bounded by SGD (CI-$\beta = 0$) and MOM (CI-$\beta = +\infty$) certainly.

% The threshold($\beta$) means we introduce the extra hyper-parameter.As is shown in Fig.~\ref{/fig:cifar_alexnet_cis},the difference in threshold($\beta$) selection will have a big impact on the results, and it’s difficult to find a suitable value.Actually, recently work \cite{an2018pid} which imitates PID form to introduce the $K_{d}$ term faces the same drawback. Fig.~\ref{/fig:mnist_pid}shows $K_{d}$ term selection is difficult. Although \cite{an2018pid} propose a formula to choose a initial $K_{d}$ value，the suitably final value is difficult to tune.Through our experiments, we also found that $K_{d}$ has poor generalization ability on different networks of different data sets, which means that different $K_{d}$ need to be tuned in different condition.

% drawbacks: hyperparameter & cannot outperform GD or MOM

Recall that what we expect is an optimizer with short rising time $t_1$ and small maximum overshoot $\Delta\theta_{\max}$.
As illustrated in Fig.~\ref{/fig:mom_lag} previously, the momentum-based algorithm has much shorter rising time $t_1$ than GD due to the accumulated gradients. However, the historical gradients lag the update of weights in the period $[t_1, t_2]$ when the gradient direction gets reversed, and lead to severe oscillation about the optimal point. To ease the fluctuation, the proposed SPI-Optimizer isolates the integral component of the controller when the inconsistency of current and historical gradient direction occurs, i.e.,
\begin{equation} 
\text{sgn}(\nabla L(\theta_t^{(i)})) \text{sgn}(v_t^{(i)}) = -1.
\end{equation}
The SPI-Optimizer is further presented by:
% sign consistency
\begin{equation} 
    \begin{aligned} 
        v_{t+1}^{(i)} &= \alpha v_t^{(i)} \mathds{1}\{\text{sgn}(\nabla L(\theta_t^{(i)})) \text{sgn}(v_t^{(i)})\} + r \nabla L(\theta_t^{(i)}),  
        %\theta_{t+1}^{(i)} &= \theta_t^{(i)} + v_{t+1}^{(i)},
    \end{aligned}
\end{equation}
The key insight here is that the historical gradients will lag the update of weights if the weights should not keep the previous direction, i.e., $\text{sgn}(\nabla L(\theta_t^{(i)})) \text{sgn}(v_t^{(i)}) \neq 1$, leading to oscillation of gradients about the optimal point until the gradients compensates the momentum in the reversed direction. In this way, SPI-Optimizer can converge as fast as MOM and NAG yet leads to much smaller maximum overshoot. On the other hand, we may interpret the SPI-Optimizer from the perspective of state delay.

\textbf{State Delay: }
Recall that the objective of this feedback system is to let the gradient $\nabla L(\theta_{t+1})$ approach $\textbf{0}$, yet we only have the observation of the state $\nabla L(\tilde{\theta}_t)$.
This can be understood as a dynamic system with measurement latency or temporal delay. The larger the delay is, the more likely severe oscillation or unstable system occurs. 
Analogously, we define $d_{t+1} = \lvert\theta_{t+1} - \tilde{\theta}_t\rvert$ as \textit{state delay} in stochastic optimization:
\begin{equation} \label{eq:delay}
    \begin{aligned} 
    d_{t+1}^{\text{GD}}  &= |r \nabla L(\theta_t)| \\
    d_{t+1}^{\text{MOM}} &= |\alpha v_t + r \nabla L(\theta_t)| \\
    d_{t+1}^{\text{NAG}} &= |r \nabla L(\theta_t - \alpha v_t)|\\
    \end{aligned}
\end{equation}
%The ideal case of this dynamic system control would be that we can have the observation on the objective of the feedback system, i.e., we can predict the gradient of the future weight $\nabla L(\theta_\{t+1\})$.
%However, that is impossible. have In dynamic control applications, the different 

% One hypothesis is that for the same type of controller (say, PI controller) the optimizer with smaller state delay works better.
\textbf{Hypothesis}: One hypothesis is that for the momentum-based optimizer (PI controller), the optimizer with smaller state delay is highly likely having less oscillation, which is harmful for system stability.
% state delay <--> oscillation 
As the increase of $\alpha$ in Eqn. \ref{eq:delay} from GD to MOM, state delay of MOM $d_{t+1}^{\text{MOM}}$ has higher chance to be larger than that of GD $d_{t+1}^{\text{GD}}$, which explains why MOM usually oscillates more during optimization process.
% \Ji{The convergence path on the banana function don't have much oscillation --> MOM works the best. For the quadratic function, oscillation appears --> NAG works better than MOM. GD doesn't oscillate much but convergant slowly}
Similarly, NAG can be understood as a more robust PI controller with smaller state delay under the assumption that both $\nabla L(\theta_t - \alpha v_t)$ and $\nabla L(\theta_t)$ share the same probabilistic distribution. For SPI-Optimizer, when the oscillation is detected, we reduce the state delay by assigning $d_{t+1}^{\text{SPI}} = d_{t+1}^{\text{GD}}$. Otherwise, it remains using PI controller to speed up.

\subsection{Discussion} 
\label{sec:toy}

\begin{figure}[t]
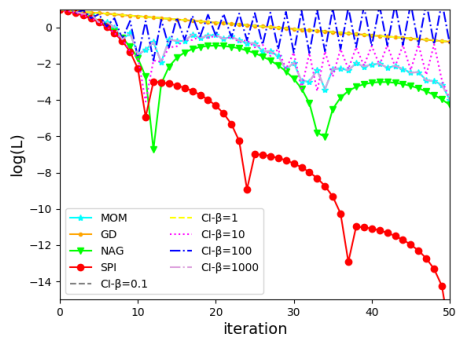
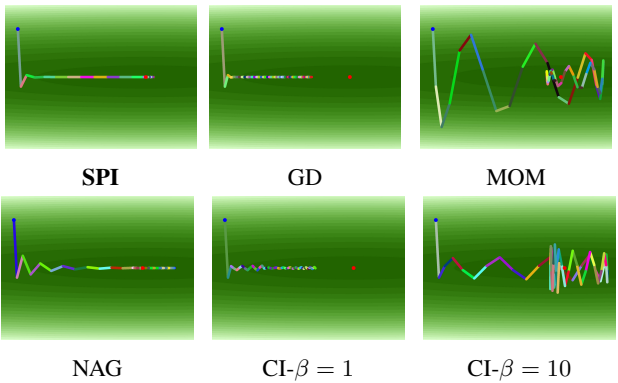
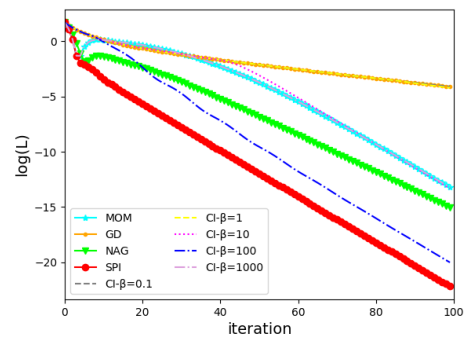
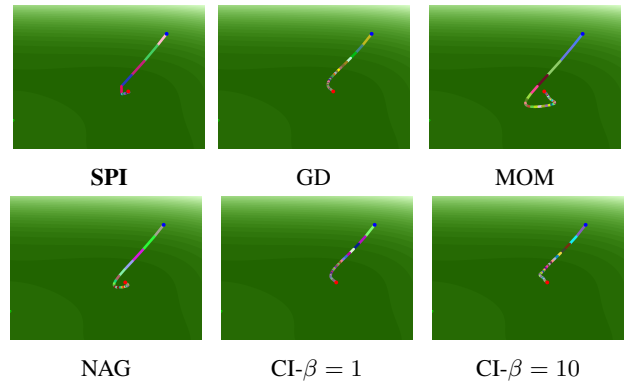

    \centering
    \newcommand{\colw}{0.305}
    \newcommand{\figw}{1.0} % can bigger than 1
    \begin{subfigure}[b]{0.8\linewidth}
        \includegraphics[width=1.0\textwidth,clip]{{{sections/figures/figure_files/function_loss_curve1}}}
        \captionsetup{labelformat=empty}
        %\caption{\textbf{SPI}}
    \end{subfigure}%
    
    \begin{subfigure}[b]{\colw\linewidth}
        \includegraphics[width=\figw\textwidth,clip]{{{sections/figures/figure_files/1_3_-2_1}}}
        \captionsetup{labelformat=empty}
         \caption{\textbf{SPI}}
    \end{subfigure}%
    ~    
    \begin{subfigure}[b]{\colw\linewidth}
        \includegraphics[width=\figw\textwidth,clip]{{{sections/figures/figure_files/1_1_-2_1}}}
        \captionsetup{labelformat=empty}
        \caption{GD}
    \end{subfigure}
    ~ 
     % leave a blank line to change row       
    \begin{subfigure}[b]{\colw\linewidth}
        \includegraphics[width=\figw\textwidth,clip]{{{sections/figures/figure_files/1_0_-2_1}}}
        \captionsetup{labelformat=empty}
        \caption{MOM}
    \end{subfigure}
    ~ 
    \begin{subfigure}[b]{\colw\linewidth}
        \includegraphics[width=\figw\textwidth,clip]{{{sections/figures/figure_files/1_2_-2_1}}}
        \captionsetup{labelformat=empty}
        \caption{NAG}
    \end{subfigure}
    ~ 
    \begin{subfigure}[b]{\colw\linewidth}
        \includegraphics[width=\figw\textwidth,clip]{{{sections/figures/figure_files/1_5_-2_1}}}
        \captionsetup{labelformat=empty}
        \caption{CI-$\beta=1$}
    \end{subfigure}
    ~ 
    \begin{subfigure}[b]{\colw\linewidth}
        \includegraphics[width=\figw\textwidth,clip]{{{sections/figures/figure_files/1_6_-2_1}}}
        \captionsetup{labelformat=empty}
        \caption{CI-$\beta=10$}
    \end{subfigure}
%    \caption{comparison of 4 popular methods on the MNIST dataset using CNN as the architecture for 20 epochs.Top row: the curves of training loss and validation loss.Bottom row: the curves of training accuracy and validation accuracy }
    \caption{Convergence comparison within 50 iterations among our SPI-Optimizer, Momentum (MOM) \cite{rumelhart1986learning}, Gradient Descent (GD) \cite{cauchy1847methode}, Nesterov Accelerated Gradient (NAG)  \cite{nesterov1983method} and conditional integral (CI-$\beta$) on the quadratic function $f_1(\theta)$.
% \ref{subfig:Nv1_weighted} to \ref{subfig:Nv6_weighted}
        } 
  \label{fig:quadratic}
\end{figure}

\begin{figure}[t]
    \centering
    \newcommand{\colw}{0.305}
    \newcommand{\figw}{1.0} % can bigger than 1
    \begin{subfigure}[b]{0.8\linewidth}
        \includegraphics[width=1.0\textwidth,clip]{{{sections/figures/figure_files/function_loss_curve13}}}
        \captionsetup{labelformat=empty}
        %\caption{\textbf{SPI}}
    \end{subfigure}%
    
    % leave a blank line to change row       
    \begin{subfigure}[b]{\colw\linewidth}
        \includegraphics[width=\figw\textwidth,clip]{{{sections/figures/figure_files/13_4_6_03}}}
        \captionsetup{labelformat=empty}
         \caption{\textbf{SPI}}
    \end{subfigure}%
    ~
    \begin{subfigure}[b]{\colw\linewidth}
        \includegraphics[width=\figw\textwidth,clip]{{{sections/figures/figure_files/13_4_6_01}}}
        \captionsetup{labelformat=empty}
        \caption{GD}
    \end{subfigure}
    ~ 
    \begin{subfigure}[b]{\colw\linewidth}
        \includegraphics[width=\figw\textwidth,clip]{{{sections/figures/figure_files/13_4_6_00}}}
        \captionsetup{labelformat=empty}
        \caption{MOM}
    \end{subfigure}
    ~ 
    \begin{subfigure}[b]{\colw\linewidth}
        \includegraphics[width=\figw\textwidth,clip]{{{sections/figures/figure_files/13_4_6_02}}}
        \captionsetup{labelformat=empty}
        \caption{NAG}
    \end{subfigure}
    ~ 
    \begin{subfigure}[b]{\colw\linewidth}
        \includegraphics[width=\figw\textwidth,clip]{{{sections/figures/figure_files/13_4_6_05}}}
        \captionsetup{labelformat=empty}
        \caption{CI-$\beta=1$}
    \end{subfigure}
    ~ 
    \begin{subfigure}[b]{\colw\linewidth}
        \includegraphics[width=\figw\textwidth,clip]{{{sections/figures/figure_files/13_4_6_06}}}
        \captionsetup{labelformat=empty}
        \caption{CI-$\beta=10$}
    \end{subfigure}
%    \caption{comparison of 4 popular methods on the MNIST dataset using CNN as the architecture for 20 epochs.Top row: the curves of training loss and validation loss.Bottom row: the curves of training accuracy and validation accuracy }
    \caption{Performance comparison within 100 iterations among our SPI-Optimizer, Momentum (MOM) \cite{rumelhart1986learning}, Gradient Descent (GD) \cite{cauchy1847methode}, Nesterov Accelerated Gradient (NAG)  \cite{nesterov1983method} and conditional integral (CI-$\beta$) on the 2D non-convex McCormick function $f_2(\theta)$.}
    
% \ref{subfig:Nv1_weighted} to \ref{subfig:Nv6_weighted}

   \label{fig:f13_McCormick}
\end{figure}

To make the hypothesis mentioned above more intuitive and rigorous, and to further quantify how much SPI-Optimizer improves compared with other optimizer, we specifically take a very simple convex function $f_1(\theta)={\theta^{(1)}}^2+50{\theta^{(2)}}^2$ and the McCormick function $f_2(\theta)=\sin(\theta^{(1)}+\theta^{(2)})+(\theta^{(1)}-\theta^{(2)})^2-1.5\theta^{(1)}+2.5\theta^{(2)}+1$ as examples to visualize the optimization procedure. The concerned representative P or PI based optimizers used for comparison are Momentum (MOM) \cite{rumelhart1986learning}, Gradient Descent (GD) \cite{robbins1951stochastic}, Nesterov Accelerated Gradient (NAG)  \cite{nesterov1983method} and Conditional Integration (CI-$\beta$) \cite{astrom1989integrator} with different thresholding parameter $\beta$.

The convergence path of each optimizer applying on the two functions are depicted in Fig.~\ref{fig:quadratic} and Fig.~\ref{fig:f13_McCormick} (sub-figures with green background). The optimal point locates at the origin $(3, 2)$ (the red point), and the convergence process starts from the blue point with the maximum 100 iterations. The loss is defined as $L=\left\Vert {f(\theta) - f(\theta^*)} \right\Vert_2$ with $r = 0.012,0.001$,respectively. Apparently, GD and SPI oscillate the least, and NAG tends to be more stable than MOM. This intuitively validates the previous hypothesis: for the same type of optimization controller, the one with smaller state delay is highly likely having less oscillation. 

Additionally, the convergence speed of all the methods can be inferred from the top chart in Fig.~\ref{fig:quadratic} and Fig.~\ref{fig:f13_McCormick}, where the naive conditional integration inspired controller with different thresholds $\beta$ are marked as CI-$\beta=\{0.1, 1, 10, 100, 1000\}$.
From the definition of CI-$\beta$, we can tell that the performance of CI-$\beta$ is almost bounded by GD (CI-$\beta = 0$) and MOM (CI-$\beta = +\infty$), which can also be interpreted from both Fig.~\ref{fig:quadratic} and Fig.~\ref{fig:f13_McCormick}. It is worthy note that the hyper-parameter $\beta$ aggravates the parameter tuning methodology, since it should be determined by the characteristics of the loss functions $L$ that depends on the training data $\left\langle x,y \right\rangle$, the network structure $f(\cdot|\theta)$ and the metric between $y = f(x|\theta^*)$ and $y = f(x|\theta)$.
Even in the toy 2D example, the extra introduced hyper-parameter $\beta$ by the CI-$\beta$ is not reliable for a favorable result.
% The final convergence epoch are  SPI:63, MOM:272, SGD:531, NAG:162, CI-$\beta=0.1$:more than 1000, CI-$\beta=1000$:272 (implying the disable of integral separation, which is a MOM) for the convex function, and SPI:61, MOM:201, SGD:568, NAG:129, CI-$\beta$=0.1:242, CI-$\beta$=1000:201 for the non-convex function. 

In contrast, the proposed SPI-Optimizer takes precautions against oscillation that may lead to unstable system and slow convergence, by preventing large state delays. So that the fluctuation phenomenon of the convergence curve gets eased.
Meanwhile, the convergence rate of SPI is clearly superior to that of others, not only in the initial stages where the value of error function is significant, but also in the later part when the error function is close to the minimum.
%More interestingly, the integral separated PI controller can outperform both P (GD) and PI (MOM) controller.
Quantitatively the convergence speed reaches up to $8\%$ and $33\%$ epoch reduction ratio respectively on the 2D function $f_1(\theta)$ and $f_2(\theta)$ when the L2 norm of the residual hits the threshold 1e-5.
%That illustrates the novelty of the proposed SPI to some extent.

% This scheme removes excessive oscillations and at the same time retains the advantages of the integral term. This can be clearly seen from

{\bf Convergence analysis:} More importantly, we conduct theoretical analysis on the convergence of SPI-Optimizer, and show that under certain condition that the loss function $f(\theta)$ is $\mu$-strongly convex and $L$-smooth\cite{richtarik2017stochastic}, and the learning rate $r$ and momentum parameter $\alpha$ within a proper range, 1) the convergence of SPI-Optimizer can be guaranteed strictly, 2) the convergence rate of SPI-Optimizer is faster than MOM. Due to limited space, the detailed analysis is presented in the supplementary material.

%\input{sections/method_convergence_proof.tex} 

% Mengqi's notes
% An example is presented now to demonstrate the effect of SPI
% exp: (in)consistence sign
    %Consequently, SPI can converge faster than both CI and classical PI controller, such as MOM and NAG when the PI controller introduce severe oscillation.
    % \Ji{The convergence path on the banana function don't have much oscillation --> MOM works the best. For the quadratic function, oscillation appears --> NAG works better than MOM. GD doesn't oscillate much but convergant slowly}

% * full CIFAR 100 show: more robust / higher acc. / faster conv.
% * generalization: compare with state-of-the-art

% # suppli:
% * show conv curve in [supplimentary material]
% * selected NO.s from CIFAR 10 & MNIST

\section{Experiments}
Following the discussion on the 2D demo with harsh fluctuation, this section studies the performance of SPI-Optimizer on the convolutional neural networks (CNNs), consisting of convolution layers and non-linear units.

In the first subsection, we compare with the most relevant solutions for dealing with the oscillation problem of the integral component. 
One method is the PID-$K_d$ \cite{an2018pid} that adds a derivative (D) term to MOM. Another counterpart is the conditional integration (CI-$\beta$) optimizer, introducing a hyper-parameter $\beta$ to define a bound within which the momentum term gets prevented. In contrast, the proposed SPI-Optimizer does not introduce extra hyper-parameter and outperforms both of them.

Subsequently, experiments are conducted to evaluate the P controller (SGD) and the PI controllers (MOM and NAG) under various training settings, showing that SPI-Optimizer is more robust to large learning rate range, different learning rate decay schemes, and various network structures.

Finally, SPI-Optimizer is compared with the state-of-the-art optimization methods in different datasets across different network structures to illustrate its better generalization ability.

Note that all the reported charts and numbers are averaged after 3 runs.

%首先用2D函数表明我们算法收敛快,震荡少的特性.由于2D函数表现好，所以看看CNN。首先在MNIST数据集上比较PID。然后在CIFAR上比较其他算法。最后比较算法表和网络表,验证泛化能力。

\begin{figure}[h!]
\centering
\includegraphics[width=0.45\textwidth]{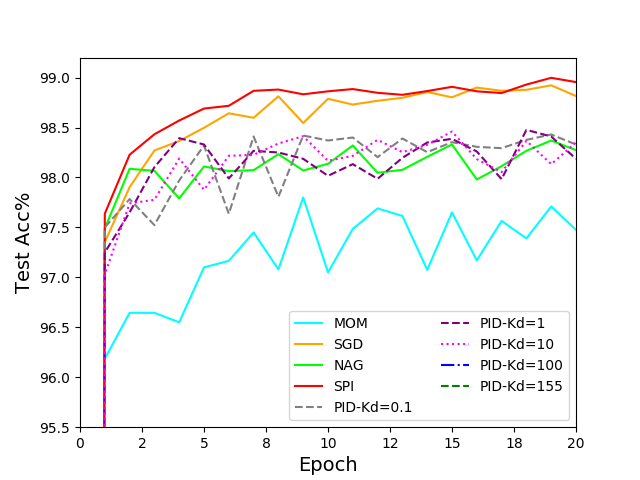} % don't need to specify .png or .jpg
\caption{Performance comparison between SPI-Optimizer and PID-$K_d$ with different $K_d$ on the MNIST dataset. Compared with SPI-Optimizer without extra hyper-parameter, PID-$K_d$ requires big effort to be tuned.}
\label{/fig:mnist_pid}
\end{figure}

\subsection{Compare with the Oscillation Compensators}
% MNIST show CI, PID (different d)
% why we don't trust D controller
% tune d --> not good results

\textbf{Comparison with PID-$K_d$}: 
%We implement various optimizers on the same CNN\footnote{The architecture of CNN contains 2 alternating stages of 5x5 convolution lifters and 2x2 max pooling with stride of 1 followed by a fully connected layer output. The dropout noise applies on the fully connected layer.}, including our SPI-Optimizer, the latest PID-Optimizer \cite{an2018pid} (PID-$K_d$), SGD-Momentum (MOM) \cite{rumelhart1986learning}, Stochastic Gradient Descent (SGD) \cite{robbins1951stochastic}, Nesterov Accelerated Gradient (NAG)  \cite{nesterov1983method}.
Fig.~\ref{/fig:mnist_pid} depicts the performance using CNNs\footnote{The architecture of CNNs contains 2 alternating stages of 5x5 convolution lifters and 2x2 max pooling with stride of 1 followed by a fully connected layer output. The dropout noise applies on the fully connected layer.} on the handwritten digits dataset MNIST \cite{lecun1998gradient} consisting of 60k training and 10k test 28x28 gray images. Even though PID-$K_d$ \cite{an2018pid} can ease the oscillation problem of MOM, its hyper-parameter $K_d$ requires much effort for empirical tuning to get relatively better result. Specifically, a large range of $K_d$ is tested from $0.1$ to $155$; however, SPI-Optimizer performs better in terms of faster convergence speed $74\%$ and around $33\%$ error reduction ratio than PID-$K_d$. %\wang{we choose the best performance $K_d$=1, 74\% faster ；33\% error reduction}/

One may notice that the curve for $K_d=100$ and $K_d=155$ (blue dashed line) did not appear in the Fig.~\ref{/fig:mnist_pid}, since $K_d=100$ and $K_d=155$ can not lead to convergence but $K_d=155$ is followed equation of $K_d$ initial value selection in \cite{an2018pid} when set learning rate is 0.12. It's worth pointing out that a similar situation exists for other learning rate. That can be explained by the fact that the hyper-parameter $K_d$ requires big effort to be tuned. One guess is that $K_d$ should depend on many factors, such as the training data, the network structure and the network loss metric.

Additionally, the comparison of the generalization ability across various network structures and datasets is listed in Tab.~\ref{tab:state_of_the_art}, where SPI-Optimizer also constantly outperforms PID-$K_d$. % P(SGD) /PI (MOM, NAG) / PID-$K_d$ 
More importantly, SPI-Optimizer does not introduce extra hyper-parameter.

\begin{figure}[h!]
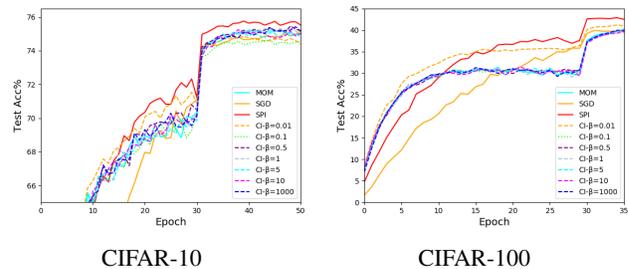

\centering

\newcommand{\colw}{0.485}
    \newcommand{\figw}{1.1} % can bigger than 1
    \begin{subfigure}[b]{\colw\linewidth}
        \includegraphics[width=\figw\textwidth,clip]{{{sections/figures/figure_files/c10yz}}}
        \captionsetup{labelformat=empty}
        \caption{CIFAR-10}
    \end{subfigure}
    ~ 
    \begin{subfigure}[b]{\colw\linewidth}
        \includegraphics[width=\figw\textwidth,clip]{{{sections/figures/figure_files/c100yz}}}
        \captionsetup{labelformat=empty}
        \caption{CIFAR-100}
    \end{subfigure}

\caption{Comparison between Condition Integration (CI-$\beta$) and SPI-Optimizer in CIFAR dataset using AlexNet. The performance of CI-$\beta$ is almost bounded by SGD (CI-$\beta = 0$) and MOM (CI-$\beta = +\infty$). SPI-Optimizer outperforms all of them by a large margin without introducing hyper-parameter.}
\label{/fig:cifar_alexnet_cis}
\end{figure}

\begin{table*}[ht]
\centering
\small
%\hspace{-0.5cm}
\begin{tabular}{|c|c|c|c|c|c|c|}
\hline
\multirow{2}{*}{Methods} & \multicolumn{3}{c|}{CIFAR-10} & \multicolumn{3}{c|}{CIFAR-100} \\ \cline{2-7} 
    &      lr=0.05 	        &       lr=0.1  	      &    lr=0.18           &   lr=0.05 	        &       lr=0.1  	      &    lr=0.18    \\ \hline
SGD &      24.812\%        	&       24.757\%	      &    25.522\%          &      60.089\%	    &    60.240\%   &   60.079\%    \\ \hline
MOM &      24.929\%	        &       27.766\%	      &    NaN                 &      59.158\%        &    68.312\%   &    NaN   \\ \hline
NAG &      24.753\%         &       26.811\%	      &    NaN                 &      58.945\%	    &  67.091\%     &  NaN    \\ \hline
\textbf{SPI} &      \textbf{24.257\%}&       \textbf{24.245\%} &    \textbf{25.823\%} &   \textbf{58.188\%}  &   \textbf{57.179\%}    &  \textbf{57.223\%}    \\\hline
\end{tabular}
\captionof{table}{
Three learning rate values are evaluated on both CIFAR-10 and CIFAR-100 datasets with AlexNet. Test errors are listed. Note that the symbol “NaN” indicates that the optimization procedure cannot converge with that specific setting. 
Compared with other P/PI optimizers (SGD, MOM, and NAG), SPI-Optimizer is more robust to larger learning rate while retaining the best performance.
} \label{tab:cifar_alexnet_diff_lr}
\end{table*}

\textbf{Comparison with CI-$\beta$}:
%Considering the impact of the hyper-parameter problem of PID-$K_d$, we propose SPI-Optimizer which inspired by the classical conditional integration controller (CI-$\beta$), but without introduction of any extra hypr-parameter .The implementations of CI-$\beta$ on CIFAR-10 and CIFAR-100 are shown in Fig.~\ref{/fig:cifar_alexnet_cis}.
%Sec. ~\ref{sec:toy}
As we observe from the toy examples in Section 3.3, the performance of CI-$\beta$ is almost bounded by SGD (CI-$\beta = 0$) and MOM (CI-$\beta = +\infty$).
%It is favorable to verify the argument in Sec. \cite{sec_CI}, that the performance of CI-$\beta$ is almost bounded by GD ($\beta = 0$) and MOM ($\beta = +\infty$). 
Similarly, from Fig.~\ref{/fig:cifar_alexnet_cis} we get the same conclusion that CI-$\beta$ can hardly outperforms SGD and MOM in a large searching range of $\beta$.
The comparison is conducted on the larger datasets CIFAR-10 \cite{krizhevsky2009learning}\footnote{The CIFAR-10 dataset consists of 60000 32x32 colour images in 10 classes, with 6000 images per class. There are 50000 training images and 10000 test images.} and CIFAR-100 \cite{krizhevsky2009learning}\footnote{The CIFAR-100 is just like the CIFAR-10, except it has 100 classes containing 600 images each.} using AlexNet \cite{krizhevsky2012imagenet}.
Quantitatively, without any extra hyper-parameter, the proposed SPI-Optimizer can reach higher accuracy ($3\%$ error reduction ratio) and faster convergence ($40\%$ speed up) than CI-$\beta$ with $\beta$ ranging from $0$ to $+\infty$.
%\wang{C10:ERROR1.3\%, SPEED30\%  C100:ERROR 3\% SPEED 40\%}

\subsection{Comparison with P/PI Optimizers}
% different lr works good: table AlexNet CIFAR-10/100
% (reducing) learning rate works file: figure AlexNet CIFAR-100
% same observation: CI-\beta does not work: 
% generalize to WRN ?

\begin{figure}[h!]
\centering

\newcommand{\colw}{0.48}
    \newcommand{\figw}{1} % can bigger than 1
    \begin{subfigure}[b]{\colw\linewidth}
        \includegraphics[width=\figw\textwidth,clip]{{{sections/figures/figure_files/c100_fixlr}}}
        \captionsetup{labelformat=empty}
        \caption{a: Fixed learning rate $r = 0.05$}
        \label{/fig:subfig_fixed_lr}
    \end{subfigure}
    ~
    \begin{subfigure}[b]{\colw\linewidth}
        \includegraphics[width=\figw\textwidth,clip]{{{sections/figures/figure_files/c100_alexnet_optimizer}}}
        \captionsetup{labelformat=empty}
        \caption{b: Learning rate decay}
        \label{/fig:subfig_lr_decay}
    \end{subfigure}

\caption{Trained on the CIFAR-100 dataset using AlexNet, SPI-Optimizer achieves good performance diffent learning rate decay schemes. The horizontal dotted-line corresponds to the highest accuracy of SGD, and the convergence speedup ratio of SPI-Optimizer is around $35\%$.}
\label{/fig:c100_varying_lr}
\end{figure} 
\begin{figure}[h!]
\centering

\newcommand{\colw}{0.48}
    \newcommand{\figw}{1} % can bigger than 1
    % \begin{subfigure}[b]{\colw\linewidth}
    %     \includegraphics[width=\figw\textwidth,clip]{{{sections/figures/figure_files/c10resnet56_60loss}}}
    %     \captionsetup{labelformat=empty}
    %     \caption{CIFAR10 ResNet loss}
    % \end{subfigure}
    % ~ 
    % \begin{subfigure}[b]{\colw\linewidth}
    %     \includegraphics[width=\figw\textwidth,clip]{{{sections/figures/figure_files/c100resnet56_60loss}}}
    %     \captionsetup{labelformat=empty}
    %     \caption{CIFAR100 ResNet loss}
    % \end{subfigure}
    % ~ 
    \begin{subfigure}[b]{\colw\linewidth}
        \includegraphics[width=\figw\textwidth,clip]{{{sections/figures/figure_files/c10resnet56_60acc}}}
        \captionsetup{labelformat=empty}
        \caption{CIFAR10 ResNet}
    \end{subfigure}
        ~ 
    \begin{subfigure}[b]{\colw\linewidth}
        \includegraphics[width=\figw\textwidth,clip]{{{sections/figures/figure_files/c100resnet56acc}}}
        \captionsetup{labelformat=empty}
        \caption{CIFAR100 ResNet}
    \end{subfigure}
         ~ 
    \begin{subfigure}[b]{\colw\linewidth}
        \includegraphics[width=\figw\textwidth]{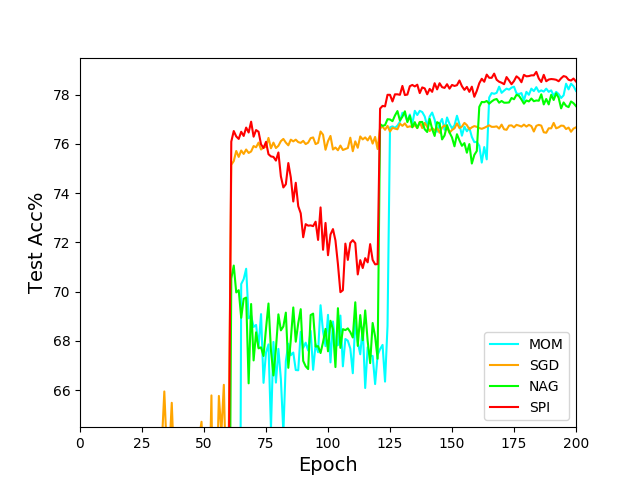} % don't need to specify .png or .jpg
        \caption{CIFAR-100 Wide-ResNet}
        \label{/fig:c100_wrn}
    \end{subfigure}
    
\caption{Trained on the CIFAR-10, CIFAR-100 dataset using ResNet and Wide-ResNet. SPI-Optimizer constantly performs the best in terms of $40\%$ faster convergence and $5\%$ error reduction ratio than the second best method.}
\label{/fig:resnet56}
\end{figure}
% \input{sections/figures/c100_wrn.tex}
% large lr
\textbf{High Adaptability Of Learning Rate}: 
The comparison for different learning rate is demonstrated in Tab.~\ref{tab:cifar_alexnet_diff_lr}, where three learning rate values are evaluated on both CIFAR-10 and CIFAR-100 datasets with AlexNet.
Note that the symbol ``NaN'' indicates that the optimization procedure cannot converge with that specific setting.
Interestingly, SPI-Optimizer is the only surviving PI controller (momentum related optimizer) in the case with large learning rate $r = 0.18$.
We can safely conclude that, compared with other P/PI optimizers (SGD, MOM, and NAG), SPI-Optimizer is more robust to larger learning rate while retaining the best performance.

\textbf{Learning Rate Decay}: 
We investigate the influence of different learning rate update schemes to the performance.
Firstly, Fig.~\ref{/fig:subfig_fixed_lr} depicts the result of constant learned rate $r=0.05$, which is corresponding to the best performance of the other methods in Tab.~\ref{tab:cifar_alexnet_diff_lr}. Even though we choose the setting with small performance gap with the others, the convergence speedup can still reach $35\%$ by comparing with the 2nd best method (SGD). It is calculated based on the (epoch, error percentage) key points of SPI-Optimizer ($125, 78\%$) and SGD ($200, 78\%$). %\wang{算ACC和SPEED应该是4个点：这样只能算speed} located by the horizontal and vertical dotted lines.

Then, as a comparison, results with decayed learning rate by a factor of 0.1 in every 30 epochs is reported in Fig.~\ref{/fig:subfig_lr_decay}.
Even though MOM and NAG rise faster in the very beginning, SPI-Optimizer still has a big accuracy improvement versus the others.
So that the proposed method can performs good in different learning rate decay schemes.
%Similarly, we show the result on Alexnet in Fig.~\ref{/fig:c100_varying_lr}.We train it for 50 epochs and the learning rate update scheme decays learning rate by a factor of 0.1 by similar learning rate  decay ie. 30 epochs .As is shown in Fig.~\ref{/fig:c100_varying_lr} (left), SPI- Optimizer was already convergent at 33 epoch, and  then it even began to overfit after this. For fair ,we choose the better initial learning rate 0.05 to each optimizer according according to  Tab.~\ref{tab:cifar_alexnet_diff_lr}.(这句话我表达的很怪,求改)
%As expected, our training process with learning rate decay is much faster than the others.
%To quantitatively measure the convergence speed, we fixed the learning rate and get around \wang{35\%} speed up at training stage compared with the best result of other methods until epoch 50, as shown in Fig.~\ref{/fig:c100_varying_lr} (right). 
%In both cases, the proposed SPI-Optimizer can perform better than the other optimizers.

\begin{table*}[tb]
\centering
\small
%\hspace{-0.5cm}
\begin{tabular}{|c|c|c|c|c|c||c|}
\hline
Optimization Methods  & \begin{tabular}{@{}c@{}}MNIST \\ \end{tabular}
        & \begin{tabular}{@{}c@{}}C10(AlexNet) \\\end{tabular}
                & \begin{tabular}{@{}c@{}}C10(WRN) \\\end{tabular}
        & \begin{tabular}{@{}c@{}}C100(AlexNet) \\\end{tabular}
                & \begin{tabular}{@{}c@{}}C100(WRN) \\\end{tabular}
                & mean error reduction ratio\\
      %  & \begin{tabular}{@{}c@{}}C10(Cliquenet) \\ \end{tabular}
             %   & \begin{tabular}{@{}c@{}}C100(Cliquenet) \\\end{tabular}\\
%    \hline
%    MOM &0.574	&0.284    &\textbf{0.627}	 \\
 %   \hline
 %   SGD      &      0.448	&\textbf{0.234}	&0.706	\\
   \hline
    SGD \cite{robbins1951stochastic}     &   1.111\%	&24.757\%	&5.252\%&60.079\%&23.392\% &7.9\%	\\
     \hline
    MOM \cite{rumelhart1986learning}    &   1.720\%	&24.929\%	&4.804\% &59.158\% &21.684\% & 11.5\%	\\
    \hline
    NAG \cite{nesterov1983method} &1.338\%	&24.753\%	&4.780\%&   58.945\%	&22.414\% & 8.3\%\\
    \hline
    Adam \cite{kingma2014adam} &1.110\%	&27.031\%	&10.254\%&  63.397\%	&32.548\% & 23.5\%\\
    \hline
    RMSprop \cite{tieleman2012lecture} &1.097\%	&30.634\%	&11.377\%&   65.704\%	&34.182\% & \textbf{27.5\%}\\
    \hline
    PID \cite{an2018pid}     &   1.3\%	&24.672\%	&5.055\%&58.946\%&21.93\% & 8.3\%	\\
     \hline
    Addsign \cite{bello2017neurall} &1.237\%  &24.811\%    &7.6\%&   60.482\%   &25.344\%  &   16.4\%\\
    \hline
    \textbf{SPI} & \textbf{ 1.070\%} &\textbf{ 24.245\% }   &\textbf{4.320}\%&\textbf{57.118\%}&\textbf{20.890}\%  & -\\
    \hline
\end{tabular}
%\hspace{0.3cm}
% put caption below the \end{tabular} to put the caption at the bottom
\captionof{table}{
Test error $e_{\text{method}}$ of the state-of-the-art methods on MNIST, CIFAR-10 and CIFAR-100 with different network structures. The mean error reduction (up to $27.5\%$) colomn averages the error reduction ratio $(e_{\text{others}} - e_{\text{SPI}}) / e_{\text{others}}$ across different network and dataset (along each row) w.r.t. the proposed method.}
\label{tab:state_of_the_art}
\end{table*}
% \hline
% IS-T=0.1 \\ $\beta=6$ $\gamma=80\%$ \\ w/o weighted average &  0.448 &   0.251 &       0.798 & 0.228 \\
 
%\input{sections/tables/networks.tex} 

\textbf{Convergence speed and Accuracy}: 
We investigate SPI-Optimizer with other optimizers on CIFAR-10 and CIFAR-100 dataset with Resnet-56\cite{he2016deep} and Wide ResNet (WRN) \cite{zagoruyko2016wide}.
\cite{he2016deep} trained ResNet-56 on CIFAR-10, CIFAR-100 by droping the learning rate by 0.1 at $50\%$ and $75\%$ of the training procedure and using weight decay of 0.0001. We use the same setting for the experiments in Fig.~\ref{/fig:resnet56}. 
WRN-16-8 is selected that consists 16 layers with widening factor $k = 8$. Following \cite{zagoruyko2016wide} training method, we also used weight decay of 0.0005, minibatch size to 128. The learning rate is dropped by a factor 0.2 at 60th, 120th, and 160th epoch with total budget of 200 epochs.
For each optimizer we report the best test accuracy out of 7 different learning rate settings ranging from 0.05 to 0.4. From Fig.~\ref{/fig:resnet56} we can see that SPI-Optimizer constantly performs the best in terms of $40\%$ faster and $5\%$ more error reduction ratio than the second best method.
%\wang{resnet c10 ,5.55\%error reduction,40\% speed} 
%\wang{resnet c100 ,4.63\% error, 45\%speed}
%\wang{WRN C100 ,3.6\%，12\%,}
%\wang{不过WRN已经在最后和原始paper比了（见文末注释）}

\subsection{Comparison with the State-of-the-art}  \label{state_of_the_art}

To further demonstrate the effectiveness and the efficiency of SPI-Optimizer, we conduct the comparison with several state-of-the-art optimizers on different datasets MNIST, CIFAR-10, and CIFAR-100 by using AlexNet and WRN, as shown in Tab.~\ref{tab:state_of_the_art}. 
The compared methods includes P controller (SGD), PI controller (MOM and NAG), Adam \cite{kingma2014adam}, RMSprop \cite{tieleman2012lecture}, PID-$K_d$ \cite{an2018pid}, and Addsign \cite{bello2017neurall},
%\wang{With WRN,SPI-Optimizer achieves error of 4.32\% in CIFAR-10 and error of 20.89\% in CIFAR-100 matching the original results by
%WRN(4.81\% and 22.07\%, respectively).}
, of which the test error $e_{\text{method}}$ is reported in the table. Additionally, the average of the error reduction ratio  ($(e_{\text{others}} - e_{\text{SPI}}) / e_{\text{others}}$) across different network and dataset (along each row) w.r.t. the proposed method is listed in the last column.
Similar conclusions as the ones in previous subsections can be made, that SPI-Optimizer outperforms the state-of-the-art optimizers by a large margin in terms of faster convergence speed (up to $40\%$ epochs reduction ratio) and more accurate classification result (up to $27.5\%$ mean error reduction ratio).
Such performance gain can also verify the generalization ability of SPI-Optimizer across different datasets and different networks.

\section{Conclusion}

By analyzing the oscillation effect of the momentum-based optimizer, we know that the lag effect of the accumulated gradients can lead to large maximum overshoot and long settling time. Inspired by the recent work in associating stochastic optimization with classical PID control theory, we propose a novel SPI-Optimizer that can be interpreted as a type of conditional integral PI controller, which prevents the integral / momentum term by examining the sign consistency between residual and integral term. Such adaptability further guarantees the generalization of optimizer on various networks and datasets. The extensive experiments on MNIST, CIFAR-10 and CIFAR-100 using various popular networks fully support the superior performance of SPI-Optimizer, leading to considerably faster convergence speed (up to $40\%$ epochs reduction ratio) and more accurate result (up to $27.5\%$ error reduction ratio) than the classical optimizers such as MOM, SGD, NAG on different networks and datasets.

%\begin{bibunit}%bib

{\small
\bibliographystyle{ieee}
\bibliography{egbib}
}

%%\putbib[egbib]
%\end{bibunit}

\newpage
\onecolumn
%\appendixpage

\centerline{\Large{Supplementary Material for the Paper:}}	
\centerline{\Large{``SPI-Optimizer: an integral-Separated PI Controller for Stochastic Optimization''}}
\vspace*{1cm}

% sections
% If you want to compile the main.tex, please comment the 1st line of the main.tex
%\newpage

%Both cases are associated with PI controller with only the difference in the dynamic system's ``observability'' on the history states $\{\tilde{\theta}_i\}_{i=0,...,t}$ and their gradients $\{\nabla L(\theta_i)\}_{i=0,...,t}$, but with the same ``controllbility'' on the state update $\Delta \theta_{t+1} = \alpha v_t - r \nabla L(\tilde{\theta}_t)$ in the current iteration, i.e., both of them calculate the weight update yet based on distinct weight states which the gradients are calculated with respect to.

\setcounter{section}{0}
\section{More 2D Examples}

%banana function
Aiming for an intuitive illustration of the performance of proposed SPI-Optimizer, we present more 2D examples on several well-known functions in optimization community, the Rosenbrock function Eqn. \ref{eq:Rosenbrock}, the Goldstein-Price function Eqn. \ref{eq:Goldstein_Price}, and a non-convex function Eqn. \ref{eq:cos}.

\begin{figure*}[t]
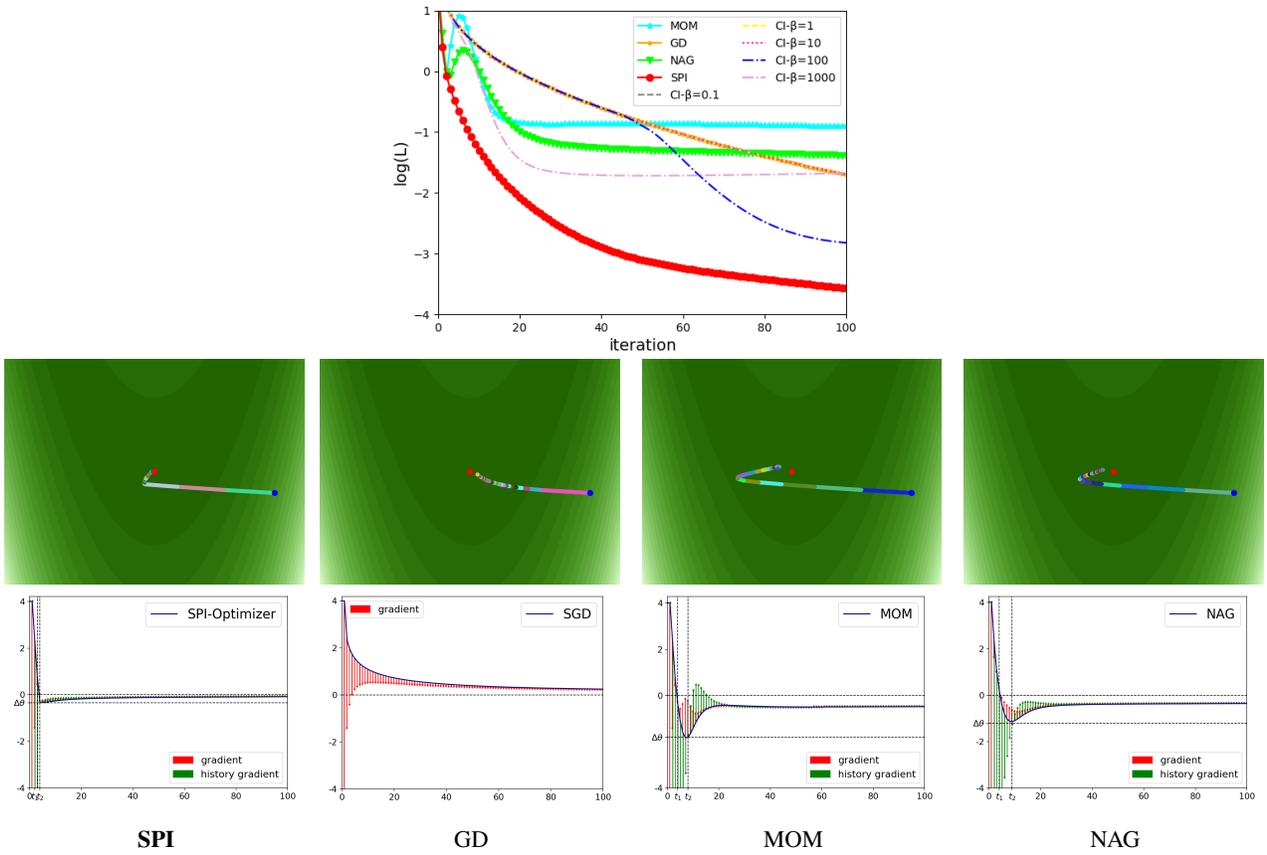

    \centering
    \newcommand{\colw}{0.23}
    \newcommand{\figw}{0.99} % can bigger than 1
    \begin{subfigure}[b]{0.40\linewidth}
        \includegraphics[width=1.0\textwidth,clip]{{{sections/figures/figure_files/function_loss_curve5}}}
        \captionsetup{labelformat=empty}
    \end{subfigure}
    
    % leave a blank line to change row    
    \begin{subfigure}[b]{\colw\linewidth}
        \includegraphics[width=\figw\textwidth,clip]{{{sections/figures/figure_files/05_4_-1.5_03}}}
        
        \includegraphics[width=\figw\textwidth,clip]{{{sections/figures/figure_files/f5overm99}}}
        \captionsetup{labelformat=empty}
         \caption{\textbf{SPI}}
    \end{subfigure}%
~    
    \begin{subfigure}[b]{\colw\linewidth}
        \includegraphics[width=\figw\textwidth,clip]{{{sections/figures/figure_files/05_4_-1.5_01}}}
        
        \includegraphics[width=\figw\textwidth,clip]{{{sections/figures/figure_files/f5oversgd}}}
        \captionsetup{labelformat=empty}
        \caption{GD}
    \end{subfigure}
~    
    \begin{subfigure}[b]{\colw\linewidth}
        \includegraphics[width=\figw\textwidth,clip]{{{sections/figures/figure_files/05_4_-1.5_00}}}
        
        \includegraphics[width=\figw\textwidth,clip]{{{sections/figures/figure_files/f5overmom}}}
        \captionsetup{labelformat=empty}
        \caption{MOM}
    \end{subfigure}
~    
    \begin{subfigure}[b]{\colw\linewidth}
        \includegraphics[width=\figw\textwidth,clip]{{{sections/figures/figure_files/05_4_-1.5_02}}}
        
        \includegraphics[width=\figw\textwidth,clip]{{{sections/figures/figure_files/f5overnag}}}
        \captionsetup{labelformat=empty}
        \caption{NAG}
    \end{subfigure}
    \caption{Convergence comparison among the proposed SPI-Optimizer, Gradient Descent (GD) \cite{cauchy1847methode}, Momentum (MOM) \cite{rumelhart1986learning}, and Nesterov Accelerated Gradient (NAG) \cite{nesterov1983method} on the Rosenbrock function Eqn. \ref{eq:Rosenbrock}.
    The \textbf{top} chart depicts the L2 distance to the optimal point in log scale over epochs.
    The \textbf{middle} row of subfigures illustrate the convergance path of the each algorithm.
    The \textbf{bottom} row of subfigures show the change of the horizontal residual w.r.t. the optimal point over epochs. Two components for the weight update are the current gradient (red arrow) and the momentum (green arrow), which can be interpreted as two forces dragging the blue curve.
        } 
   \label{fig:Rosenbrock}
\end{figure*}

Recall that the loss is defined as $L=\left\Vert {f(\theta) - f(\theta^*)} \right\Vert_2$. The top charts of Fig.~\ref{fig:Rosenbrock}, Fig.~\ref{fig:Goldstein_Price}, and Fig.~\ref{fig:cos} depict the loss $L$ in log scale over epochs. The left column of subfigures illustrates the convergence path of each algorithm. The right column of subfigures shows the change of the horizontal residue w.r.t. the optimal point over epochs. The weight update consists of the current gradient (red arrow) and the momentum (green arrow), which can be interpreted as two forces dragging the residual curve (blue).

\textbf{The Rosenbrock function}:

\begin{equation}
f_3(\theta)=(1-\theta^{(1)})^2+100(\theta^{(2)}-{\theta^{(1)}}^2)^2
\label{eq:Rosenbrock} % Eqn. \ref{eq:Rosenbrock}
\end{equation}

The convergence path depicted in Fig.~\ref{fig:Rosenbrock} starts at the point $(4,-1.5)$ (the blue dot) with the optimal point locating at $(1, 1)$ (the red dot). The lr is $r =6e-5$.

\textbf{The Goldstein-Price function}:

\begin{equation}
    \begin{split}
        f_4(\theta)= &[1+(\theta^{(1)}+\theta^{(2)}+1)^2(19-14\theta^{(1)}+3{\theta^{(1)}}^2-14\theta^{(2)}+6\theta^{(1)}\theta^{(2)}+3{\theta^{(2)}}^2)]*  \\
                     &[30+(2\theta^{(1)}-3\theta^{(2)})^2(18-32\theta^{(1)}+12{\theta^{(1)}}^2+48\theta^{(2)}-36\theta^{(1)}\theta^{(2)}+27{\theta^{(2)}}^2)]
    \end{split}
\label{eq:Goldstein_Price} % Eqn. \ref{eq:Goldstein_Price}
\end{equation}

The convergence path depicted in Fig.~\ref{fig:Goldstein_Price} starts at the point $(-4, 4.5)$ (the blue dot) with the optimal point locating at $(0, -1)$ (the red dot). The lr is $r =5e-8$.

\textbf{A 2D trigonometric function}:

\begin{equation}
f_5(\theta)= -[cos(\theta^{(1)})+1][cos(2\theta^{(2)})+1]
\label{eq:cos} % Eqn. \ref{eq:cos}
\end{equation}

The convergence path depicted in Fig.~\ref{fig:cos} starts at the point $(-2, 1)$ (the blue dot) with the optimal point locating at $(0, 0)$ (the red dot). The lr is $r = 0.012$.

Apparently, GD and SPI oscillate the least, and NAG tends to be more stable than MOM.
This intuitively validates the previous hypothesis: the optimizer with smaller state delay is less likely to oscillate around about the optimal point.

\twocolumn

\begin{figure*}[htbp]
    \centering
    \newcommand{\colw}{0.23}
    \newcommand{\figw}{0.99} % can bigger than 1
    \begin{subfigure}[b]{0.40\linewidth}
        \includegraphics[width=1.0\textwidth,clip]{{{sections/figures/figure_files/function_loss_curve11}}}
        \captionsetup{labelformat=empty}
    \end{subfigure}
    
    % leave a blank line to change row    
    \begin{subfigure}[b]{\colw\linewidth}
        \includegraphics[width=\figw\textwidth,clip]{{{sections/figures/figure_files/11_-4_4.5_03}}}
        
        \includegraphics[width=\figw\textwidth,clip]{{{sections/figures/figure_files/f11overm99}}}
        \captionsetup{labelformat=empty}
         \caption{\textbf{SPI}}
    \end{subfigure}%
 ~   
    \begin{subfigure}[b]{\colw\linewidth}
        \includegraphics[width=\figw\textwidth,clip]{{{sections/figures/figure_files/11_-4_4.5_01}}}
        
        \includegraphics[width=\figw\textwidth,clip]{{{sections/figures/figure_files/f11oversgd}}}
        \captionsetup{labelformat=empty}
        \caption{GD}
    \end{subfigure}
 ~   
    \begin{subfigure}[b]{\colw\linewidth}
        \includegraphics[width=\figw\textwidth,clip]{{{sections/figures/figure_files/11_-4_4.5_00}}}
        
        \includegraphics[width=\figw\textwidth,clip]{{{sections/figures/figure_files/f11overmom}}}
        \captionsetup{labelformat=empty}
        \caption{MOM}
    \end{subfigure}
    ~
    \begin{subfigure}[b]{\colw\linewidth}
        \includegraphics[width=\figw\textwidth,clip]{{{sections/figures/figure_files/11_-4_4.5_02}}}
        
        \includegraphics[width=\figw\textwidth,clip]{{{sections/figures/figure_files/f11overnag}}}
        \captionsetup{labelformat=empty}
        \caption{NAG}
    \end{subfigure}
    \caption{Convergence comparison among the proposed SPI-Optimizer, Gradient Descent (GD) \cite{cauchy1847methode}, Momentum (MOM) \cite{rumelhart1986learning}, and Nesterov Accelerated Gradient (NAG) \cite{nesterov1983method} on the Goldstein-Price function Eqn. \ref{eq:Goldstein_Price}.
    The \textbf{top} chart depicts the L2 distance to the optimal point in log scale over epochs.
    The \textbf{middle} row of subfigures illustrate the convergance path of the each algorithm.
    The \textbf{bottom} row of subfigures show the change of the horizontal residual w.r.t. the optimal point over epochs. Two components for the weight update are the current gradient (red arrow) and the momentum (green arrow), which can be interpreted as two forces dragging the blue curve.
        } 
   \label{fig:Goldstein_Price}
\end{figure*}

\begin{figure*}[htbp]
    \centering
    \newcommand{\colw}{0.23}
    \newcommand{\figw}{0.99} % can bigger than 1
    \begin{subfigure}[b]{0.40\linewidth}
        \includegraphics[width=1.0\textwidth,clip]{{{sections/figures/figure_files/function_loss_curve3}}}
        \captionsetup{labelformat=empty}
    \end{subfigure}
    
    % leave a blank line to change row    
    \begin{subfigure}[b]{\colw\linewidth}
        \includegraphics[width=\figw\textwidth,clip]{{{sections/figures/figure_files/3_3_-2_1}}}
        
        \includegraphics[width=\figw\textwidth,clip]{{{sections/figures/figure_files/f3overm99}}}
        \captionsetup{labelformat=empty}
         \caption{\textbf{SPI}}
    \end{subfigure}%
 ~   
    \begin{subfigure}[b]{\colw\linewidth}
        \includegraphics[width=\figw\textwidth,clip]{{{sections/figures/figure_files/3_1_-2_1}}}
        
        \includegraphics[width=\figw\textwidth,clip]{{{sections/figures/figure_files/f3oversgd}}}
        \captionsetup{labelformat=empty}
        \caption{GD}
    \end{subfigure}
 ~   
    \begin{subfigure}[b]{\colw\linewidth}
        \includegraphics[width=\figw\textwidth,clip]{{{sections/figures/figure_files/3_0_-2_1}}}
        
        \includegraphics[width=\figw\textwidth,clip]{{{sections/figures/figure_files/f3overmom}}}
        \captionsetup{labelformat=empty}
        \caption{MOM}
    \end{subfigure}
    ~
    \begin{subfigure}[b]{\colw\linewidth}
        \includegraphics[width=\figw\textwidth,clip]{{{sections/figures/figure_files/3_2_-2_1}}}
        
        \includegraphics[width=\figw\textwidth,clip]{{{sections/figures/figure_files/f3overnag}}}
        \captionsetup{labelformat=empty}
        \caption{NAG}
    \end{subfigure}
    \caption{Convergence comparison among the proposed SPI-Optimizer, Gradient Descent (GD) \cite{cauchy1847methode}, Momentum (MOM) \cite{rumelhart1986learning}, and Nesterov Accelerated Gradient (NAG) \cite{nesterov1983method} on the non-convex function Eqn. \ref{eq:cos}.
    The \textbf{top} chart depicts the L2 distance to the optimal point in log scale over epochs.
    The \textbf{middle} row of subfigures illustrate the convergance path of the each algorithm.
    The \textbf{bottom} row of subfigures show the change of the horizontal residual w.r.t. the optimal point over epochs. Two components for the weight update are the current gradient (red arrow) and the momentum (green arrow), which can be interpreted as two forces dragging the blue curve.
        } 
   \label{fig:cos}
\end{figure*}

%\input{sections/figures/trajectory_rosenbrock.tex}
%\input{sections/figures/f5_lag.tex}
%\input{sections/figures/trajectory_Goldstein_Price_function.tex}
%\input{sections/figures/f11_lag.tex}
%\input{sections/figures/trajectory_cos.tex}
%\input{sections/figures/f3_lag.tex}

% \subsection{High Adaptability Of Learning Rate in MNIST }
% \input{sections/tables/mnist_cnn.tex}
% Extend to Section 4.2, for "High Adaptability Of Learning Rate", Tab.~\ref{tab:mnist_cnn} show the SPI-Optimier's error reduction ratio on test accuracy compared with other popular methods with different learning rate. Our algorithm has some robustness compared with other algorithms ...SPI-O with different learning rate are similar,but other methods' test accuracy will decline like Tab.~\ref{tab:mnist_cnn}

% Actually,when the learning rate up to 0.15, NAG and MOM appeared non-convergence with some start points and when the learning rate up to 0.2,they can't convergence.
% However,we can know from Tab.~\ref{tab:mnist_cnn},SGD's perform still not bad.And when the learning rate up to 0.2,SGD can outperform than SPI,but SGD has the characteristics of poor final convergence which appeared on the big datasets such as CIFAR-10 and CIFAR-100 in our experiment.
% \input{sections/tables/mnist_cnn.tex}
% \subsection{State of art}

% \textbf{MNIST}:

% \input{sections/figures/e23alexnet.tex}
% \input{sections/figures/e45resnet.tex}
% \input{sections/figures/e89wrn.tex}
% \textbf{Alexnet}:
% \textbf{Resnet56}:
% \textbf{wrn}:

\onecolumn

\section{Convergence Proof of SPI-Optimizer}
Given the mathematical representation of SPI-Optimizer as
\begin{equation}
\theta_{k+1} = \theta_{k}-r \nabla f_{k} \left(\theta_{k} \right) + \alpha \left(\theta_{k}-\theta_{k-1} \right)\mathds{1}\{\text{sgn}(\nabla f_{k})\text{sgn}\left(\theta_{k-1}-\theta_{k} \right)\},
\label{eqn-SPI}
\end{equation}
we particularly introduce a diagonal matrix $\Lambda_k$ to replace the indicator function for ease of derivation. The diagonal elements in $\Lambda_k$ are all 1 or 0, indicating whether momentum terms are deserted on different dimensions. Then, we have  $\Lambda_k^{\mathrm{T}}\Lambda_k=\Lambda_k$, and Eqn. \ref{eqn-SPI} is represented as
\begin{equation}
\theta_{k+1} = \theta_{k}-r \nabla f_{k} \left(\theta_{k} \right) + \alpha \Lambda_k \left(\theta_{k}-\theta_{k-1} \right).
\end{equation}

By further denoting $\left\| \theta \right\|_{\Lambda_{k}}^2 = \theta^{\mathrm{T}} \Lambda_k \theta,\ \left\langle \theta_1, \theta_2 \right\rangle_{\Lambda_{k}}= \theta_1^{\mathrm{T}} \Lambda_k \theta_2$, and $\theta_{*}$ as the global minimum point satisfying $\nabla f(\theta_{*})=0$, it can be shown that several inequalities hold as follows,
\begin{align*}
\left\|\theta_{k}-\theta_{k-1} \right\|_{\Lambda_{k}}^{2} &= \left\|\left(\theta_{k}-\theta_{*}\right)-\left(\theta_{k-1}-\theta_{*}\right) \right\|_{\Lambda_{k}}^{2}\leq 2\left\|\theta_{k}-\theta_{*}\right\|_{\Lambda_{k}}^{2}+2\left\|\theta_{k-1}-\theta_{*}\right\|_{\Lambda_{k}}^2\\
2\left\langle \theta_{k} - \theta_ {*}, \theta_{k}-\theta_{k-1} \right\rangle_{\Lambda_{k}} &= \left\|\theta_{k} - \theta_{*}\right\|_{\Lambda_{k}}^2- \left\|\theta_{k-1} - \theta_ {*}\right\|_{\Lambda_{k}}^2+ \left\|\theta_{k} - \theta_ {k-1}\right\|_{\Lambda_{k}}^2\\
&\leq 3\left\|\theta_{k} - \theta_{*}\right\|_{\Lambda_{k}}^2+\left\|\theta_{k-1} - \theta_ {*}\right\|_{\Lambda_{k}}^2\\
2\left\langle \nabla f_k \left(\theta_{k} \right), \theta_{k-1}-\theta_{k} \right\rangle_{\Lambda_{k}}&\leq 2\left\| \nabla f_k \left(\theta_{k} \right) \right\|_{\Lambda_{k}}\left\| \theta_{k}-\theta_{k-1} \right\|_{\Lambda_{k}}\leq \left\| \nabla f_k \left(\theta_{k} \right) \right\|_{\Lambda_{k}}^2+\left\| \theta_{k}-\theta_{k-1} \right\|_{\Lambda_{k}}^2 \\
&\leq  \left\| \nabla f_k \left(\theta_{k} \right) \right\|_{\Lambda_{k}}^2+2\left\| \theta_{k}-\theta_{*} \right\|_{\Lambda_{k}}^2+2\left\| \theta_{k-1}-\theta_{*} \right\|_{\Lambda_{k}}^2
\end{align*}

Now we focus on studying whether the sequence $\{\theta_k\}$ generated by SPI converges to the best parameter point $\theta_*$ by decomposing $\left\| \theta_{k+1}-\theta_{*} \right\| ^{2}$ as follows. Note that we have combined the three inequalities above during the derivation process.
\begin{align*}
& \left\| \theta_{k+1}-\theta_{*} \right\| ^{2}\\
= & \left\| \theta_{k}-r \nabla f_{k} \left(\theta_{k} \right) + \alpha\Lambda_{k} \left(\theta_{k}-\theta_{k-1} \right)-\theta_{*} \right\|^{2} \\
= &{\left\| \theta_{k}-r \nabla f_{k} \left(\theta_ {k} \right) - \theta_{*} \right\|^{2}} + {2\alpha \left\langle \theta_{k} - r\nabla f_{k} \left(\theta_{k} \right) -\theta_ {*}, \theta_{k}-\theta_{k-1} \right\rangle_{\Lambda_{k}}} +{\alpha^2 \left\| \theta_{k}-\theta_{k-1} \right\|_{\Lambda_{k}} ^{2}} \\
= & {\left\| \theta_{k}-\theta_{*} \right\|^{2} - 2r\left\langle \nabla f_{k} \left(\theta_{k} \right), \theta_{k} - \theta_ {*}\right\rangle +r^ {2} \left\| \nabla f_k \left(\theta_{k} \right) \right\| ^{2}}\\
&+{2\alpha \left\langle \theta_{k} - \theta_ {*}, \theta_{k}-\theta_{k-1} \right\rangle_{\Lambda_{k}} + 2r\alpha\left\langle \nabla f_{k} \left(\theta_{k} \right), \theta_{k-1}-\theta_{k} \right\rangle_{\Lambda_{k}}} + {\alpha^{2} \left\|\theta_{k}-\theta_{k-1} \right\|_{\Lambda_{k}}^{2}}\\
\leq &\left\| \theta_{k}-\theta_{*} \right\|^{2} + (3\alpha+2r\alpha+2\alpha^2)\left\|\theta_{k}-\theta_{*} \right\|_{\Lambda_{k}}^{2} + (\alpha+2r\alpha+2\alpha^{2}) \left\|\theta_{k-1}-\theta_{*}\right\|_{\Lambda_{k}}^2\\
&- 2r\left\langle \nabla f_{k} \left(\theta_{k} \right), \theta_{k} - \theta_ {*}\right\rangle +r^ {2} \left\| \nabla f_k \left(\theta_{k} \right) \right\| ^{2} + r\alpha\left\| \nabla f_k \left(\theta_{k} \right) \right\|_{\Lambda_{k}}^2.
\end{align*}

Take a typical case that $f(\theta)$ is a convex function for example, which is usually assumed to be a $\mu$-strongly convex and $L$-smooth function\cite{richtarik2017stochastic}. According to the convex optimization theory, we have inequalities as follows,
\begin{align*}
\mu\text{-strongly convex}: &\left\langle \nabla f\left(\theta \right), \theta-\theta_{*} \right\rangle \geq \mu \left\|\theta -\theta_*\right\|^{2},\\
L\text{-smooth}: &\|\nabla f(\theta)-\nabla f(\theta_*)\| =\|\nabla f(\theta)\| \leq L\|\theta-\theta_*\|.
\end{align*}

While the stochastic gradients of loss function $f(\theta)$ is adopted by denoting the stochastic gradient in epoch $k$ as $\nabla f_k(\theta)$, according to the $L-$smooth property we can assume that such property holds for all stochastic gradients in the experiments, or we can increase the value of $L$ to satisfy it. Hence we have
\begin{align*}
\left\|\theta_{k+1}-\theta_{*} \right\| ^{2}\leq
&\left\| \theta_{k}-\theta_{*} \right\|^{2} + (3\alpha+2r\alpha+2\alpha^2)\left\|\theta_{k}-\theta_{*} \right\|_{\Lambda_{k}}^{2} + (\alpha+2r\alpha+2\alpha^{2}) \left\|\theta_{k-1}-\theta_{*}\right\|_{\Lambda_{k}}^2\\
&- 2r\left\langle \nabla f_{k} \left(\theta_{k} \right), \theta_{k} - \theta_ {*}\right\rangle +(r\alpha+r^2) \left\| \nabla f_k \left(\theta_{k} \right) \right\| ^{2}\\
\leq &(1+r\alpha L^2+r^2 L^2)\left\| \theta_{k}-\theta_{*} \right\|^{2} + (3\alpha+2r\alpha+2\alpha^2)\left\|\theta_{k}-\theta_{*} \right\|_{\Lambda_{k}}^{2} \\
&+ (\alpha+2r\alpha+2\alpha^{2}) \left\|\theta_{k-1}-\theta_{*}\right\|_{\Lambda_{k}}^2- 2r\left\langle \nabla f_{k} \left(\theta_{k} \right), \theta_{k} - \theta_ {*}\right\rangle.
\end{align*}
Taking expectations on both sides, we have
\begin{align*}
\mathbf{E}\left\|\theta_{k+1}-\theta_{*} \right\| ^{2}\leq &(1+r\alpha L^2+r^2 L^2)\mathbf{E}\left\| \theta_{k}-\theta_{*} \right\|^{2} + (3\alpha+2r\alpha+2\alpha^2)\mathbf{E}\left\|\theta_{k}-\theta_{*} \right\|_{\Lambda_{k}}^{2}\\
&+ (\alpha+2r\alpha+2\alpha^{2}) \mathbf{E}\left\|\theta_{k-1}-\theta_{*}\right\|_{\Lambda_{k}}^2-2r\left\langle \nabla f \left(\theta_{k} \right), \theta_{k} - \theta_ {*}\right\rangle\\
\leq & \left(1+3\alpha+2r\alpha+2\alpha^{2}+r\alpha L^2+r^2 L^2-2r\mu \right)\mathbf{E}\left\|\theta_{k}-\theta_{*} \right\| ^{2}\\
&+\left(\alpha+2r\alpha+2\alpha^{2}\right)\mathbf{E}\left\|\theta_{k-1}-\theta_{*} \right\| ^{2}.
\end{align*}
The last step above is based on the inequality $\left\| \theta \right\|_{\Lambda_{k}}^2 \leq \left\| \theta \right\|^2$. For the case of SGD-momentum, we have $\boxed{\Lambda_{k}\equiv I, \left\| \theta \right\|_{\Lambda_{k}}^2\equiv \left\| \theta \right\|^2}$. As a result, we can see that the bound of our SPI-Optimizer is more tight than that of SGD-momentum.
~\\~\\

Now we investigate whether such bound is sufficient enough for convergence. By denoting $b_k=1+3\alpha+2r\alpha+2\alpha^{2}+r\alpha L^2+r^2 L^2-2r\mu, b_{k-1}=\alpha+2r\alpha+2\alpha^{2}$, we have
\begin{align*}
\mathbf{E}\left\|\theta_{k+1}-\theta_{*} \right\| ^{2}\leq b_k\mathbf{E}\left\|\theta_{k}-\theta_{*} \right\| ^{2}+b_{k-1}\mathbf{E}\left\|\theta_{k-1}-\theta_{*} \right\| ^{2}.
\end{align*}

According to Lemma 9 in Loizou's study
\cite{loizou2017momentum}, as long as $b_{k-1}\geq 0, b_k+b_{k-1}<1$, the convergence of sequence $\{\left\|\theta_{k}-\theta_{*} \right\| ^{2}\}$ is guaranteed, which implies that
\begin{gather*}
\left(1+3\alpha+2r\alpha+2\alpha^{2}+r\alpha L^2+r^2 L^2-2r\mu \right)+\left(\alpha+2r\alpha+2\alpha^{2}\right)<1\\ \Rightarrow 4\alpha^2 + (4+4r+r L^2)\alpha+ r^2 L^2-2r\mu < 0
\end{gather*}
Considering that $r$ and $\alpha$ are positive, we have $r^2 L^2-2r\mu<0$. Hence its solution can be given as
\begin{align*}
0<r<\frac{2\mu}{L^2},\ 0<\alpha<\frac{-(4+4r+r L^2)+\sqrt{(4+4r+r L^2)^2+16(2r\mu-r^2 L^2)}}{8}.
\end{align*}

The result indicates that the convergence of our SPI-Optimizer can be guaranteed under certain values of $r$ and $\alpha$. It is worth noting that we have used the inequality $\left\| \theta \right\|_{\Lambda_{k}}^2\leq \left\| \theta \right\|^2$ during our derivation, while for the SGD-momentum algorithm, none of the components is discarded. Consequently, the bound of our SPI-Optimizer will be tighter than that of SGD-momentum. In other words, our SPI-Optimizer tends to converge faster than SGD-momentum under certain parameters.

{\small
	\bibliographystyle{ieee}
	\bibliography{egbib}
}

\end{document}